\documentclass{article}

\usepackage{PRIMEarxiv}

\usepackage[utf8]{inputenc} % allow utf-8 input
\usepackage[T1]{fontenc}    % use 8-bit T1 fonts
\usepackage{hyperref}       % hyperlinks
\usepackage{url}            % simple URL typesetting
\usepackage{booktabs}       % professional-quality tables
\usepackage{amsfonts}       % blackboard math symbols
\usepackage{nicefrac}       % compact symbols for 1/2, etc.
\usepackage{microtype}      % microtypography
\usepackage{lipsum}
\usepackage{fancyhdr}       % header
\usepackage{graphicx}       % graphics
\graphicspath{{media/}}     % organize your images and other figures under media/ folder

\title{Detection of Tool based Edited Images from Error Level Analysis and Convolutional Neural Network \\
}

\author{
  Abhishek Gupta \\
  Engineer \\
  University of Mumbai \\
  Mumbai\\
  \texttt{abhishek.gupta20001@gmail.com} \\
   \And
  Raunak Joshi \\
  Research Scholar \\
  University of Mumbai \\
  Mumbai\\
  \texttt{raunakjoshi.m@gmail.com} \\
   \And
   Ronald Laban \\
   Assistant Professor \\
   SJCEM \\
   Palghar \\
   \texttt{ronaldlaban@gmail.com} \\
}

\begin{document}
\maketitle

\begin{abstract}
Image Forgery is a problem of image forensics and its detection can be leveraged using Deep Learning. In this paper we present an approach for identification of authentic and tampered images done using image editing tools with Error Level Analysis and Convolutional Neural Network. The process is performed on CASIA ITDE v2 dataset and trained for 50 and 100 epochs respectively. The respective accuracies of the training and validation sets are represented using graphs.
\end{abstract}

% keywords can be removed
\keywords{Convolutional Neural Network \and Error Level Analysis \and Image Forensics \and Image Tampering}

\section{Introduction}
The field of Deep Learning\cite{lecun2015deep} has subsequently made groundbreaking research in various areas of vision, linguistics and audio but the basis of it can be traced back from the days of Machine Learning\cite{sah2020machine}. Basically the idea revolves around the prognostication of patterns from the data with differences. Machine Learning was certainly derived from the inferential statistic and later grew to become an independent area of research. The machine learning is derived in primary divisions of classification and regression, where we focus on classification being the problem of this paper. Binary Classification\cite{10.5120/ijca2017913083} is the precise type of classification that we want to focus on. The binary classification starts with various algorithms in field of machine learning, starting from logistic regression\cite{cramer2002origins} to discriminant analysis\cite{gupta2022discriminant}, support vectors\cite{hearst1998support}, nearest neighbors\cite{cunningham2021k}, bagging\cite{kanvinde2022binary}, boosting ensemble\cite{gupta2021succinct} and stacking generalization\cite{nair2022combining} methods, but the problem we focus on is much more in depth. The need of deep learning is evident as many state of the art machine learning algorithms we want to use are going to fall short on performance for our problem. The area of deep learning works with more samples of data and in more detailed fashion, almost replicating the human tendencies. The deep learning works with the concept of artificial neural network that replicates the biological neuron available in human brain. It is capable of deriving patterns with great depth and create quite a difference when compared with the tradition machine learning algorithms. The learning representations in done with the help of forward propagation\cite{548917} where the features are learnt by the network with the help of weights and arbitrarily declared biases. These later are given to activation functions for deeper learning and minimizing the load on the network. Later the loss is computed after completion of one forward propagation and it is optimized in the backward propagation\cite{10.5555/65669.104451} process where a loss optimizer is used that computes the gradients of the loss and retraces to weights for better learning. This entire process is considered as one epoch of training. This process is repeated several times till the deeper representation of the patterns from the data is done. The problem we focus on uses computer vision with deep learning, where a specialized neural network specially designed for vision problems is used, known as Convolutional Neural Network\cite{lecun1995convolutional}. This recognizes patterns from the images considering the number of channels. The process starts by assigning various filters in the learning process that help the network to get greater details of the image. These are termed as the features for the network which are also compressed in dimensions and increased in depth with a fashion to consider only the important features. These features are later flattened and treated as an ordinary neural network to work with the output layer. The problem that we focus on is classification of an image after using editing tools. The editing tools in the area of deep learning\cite{joshi2022refactoring} can be used but this is about detection of the edited image using deep learning. The similar concepts will be elaborated in further sections of the paper to work with problem of the forged images with editing tools.

\section{Methodology}
This section of paper focuses on the workflow design and implementation of the problem. The problem revolves around the detection of tampered image with error level analysis given to deep learning model. The dataset is the first and most important portion of the implementation.

\subsection{Dataset}
The dataset we used for this problem is CASIA dataset. The CASIA ground-truth\cite{pham2019hybrid} dataset is prominent version that contains 8 categories of images. CASIA ground-truth dataset has an extension that is being used as the tampered images dataset. Known as CASIA ITDE v.2\cite{Dong2013}, basically the dataset contains 2 prime classes, authentic images and tampered images. These tampered images are done using photoshop image editing tool. 

\subsection{Error Level Analysis}
Image forensics\cite{Piva2013AnOO} is the branch where Error Level Analysis\cite{7412439} also abbreviated as ELA is used for identification of the image portions with different compression levels is done. The representation of this can be done for detection of the tampered images using the editing tool. This can practically turn out to be a very good preprocessing step before proceeding with model. The authentic image can be observed below

\begin{figure}[htbp]
    \centering
    \includegraphics[scale=1.25]{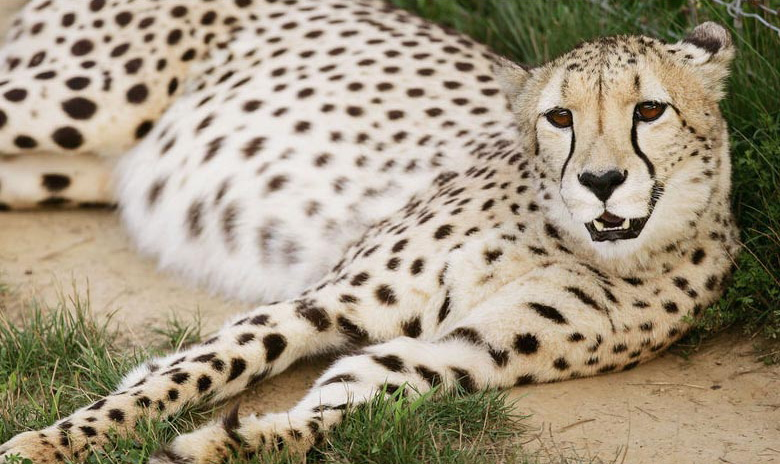}
    \caption{Authentic Image}
    \label{fig:fig1}
\end{figure}

\begin{figure}[htbp]
    \centering
    \includegraphics[scale=1.25]{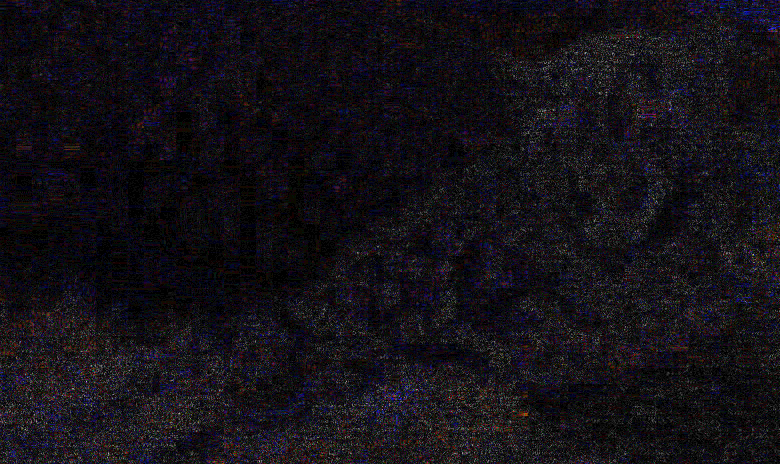}
    \caption{Authentic Image under ELA}
    \label{fig:fig2}
\end{figure}

The application of the ELA can be done. It also considers a threshold value that determines the quality of image that highlights the edges. 90\% is the quality value used which can be seen in the figure \ref{fig:fig2}.

The tampered images in similar fashion can be observed, the image and its ELA version.

\begin{figure}[htbp]
    \centering
    \includegraphics[scale=0.4]{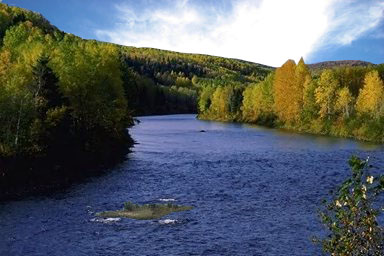}
    \caption{Tampered Image}
    \label{fig:fig3}
\end{figure}

\begin{figure}[htbp]
    \centering
    \includegraphics[scale=.4]{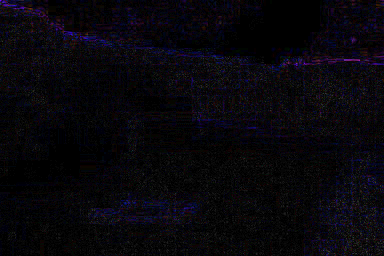}
    \caption{Tampered Image under ELA}
    \label{fig:fig4}
\end{figure}

The tampered image under ELA can be seen in figure \ref{fig:fig4}.

These images certainly point out the details from the dataset and their respective divisions. The images are used as the data for training the network.

\subsection{Network}
The network used is a Convolutional Neural Network that has many elements involved. The input layer, hidden layers and output layer are the primary subdivisions of the entire structure.

\begin{figure}[htbp]
    \centering
    \includegraphics[scale=0.9]{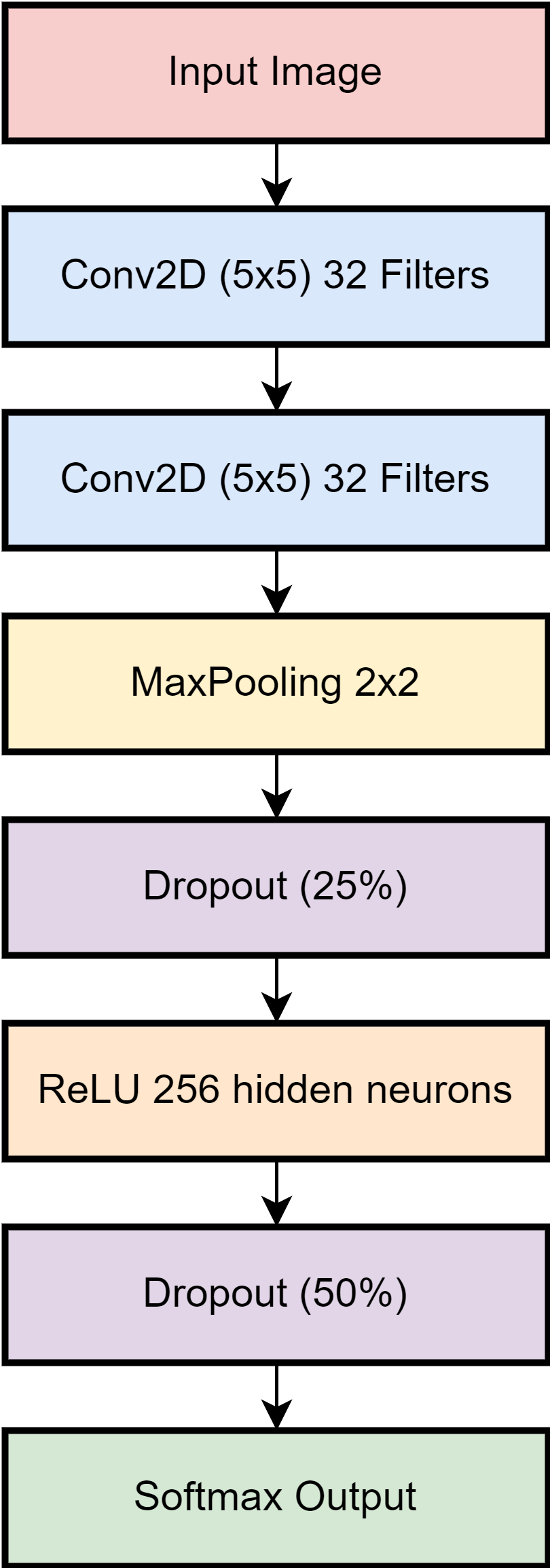}
    \caption{Network used for Implementation}
    \label{fig:fig5}
\end{figure}

The input images are given to a 2-Dimensional Convolutional Layer which is responsible for feature extraction from the image by using a set of filters. The layer uses 32, 5x5 dimensional set of filters. ReLU\cite{agarap2018deep} is the activation function used which is basically rectified linear unit. Similarly one more convolutional 2-dimensional layer with 32, 5x5 filters are used along with ReLU activation. The parameters learnt by first layer are 2432. The parameters learnt by second layer are 25632. Followed by the 2 convolutional layers, 2-dimensional max pooling\cite{gholamalinezhad2020pooling} layer is used. This specifically removes the unnecessary feature information and reduces the dimension of the activation layer. Parameters learnt from this layer are zero as main motive of the layer is to reduce the dimension. Although the dimensions after reduction keep the adequate amount of information, very subsequent amount of information can be discarded because it affects the bias of the network. Much more information when held in the process of forward propagation can be responsible for poor learning representations. Even in the later stages the network during the backward propagation can run into vanishing gradient\cite{10.1142/S0218488598000094} problem where calculating the gradients with loss optimizers becomes a daunting task. In such a scenario Dropout\cite{JMLR:v15:srivastava14a} layers can be used. They discard a lot of useless information making the networks learn better. Certain amount of information to be dropped from a neuron needs to be specified. This amount is specified in percentages. The threshold we used is 25\%,  and parameters learnt are zero. Even the dimension of the layers stay the same in this layer. Now we flatten the entire conv layers for further process. Now the process works like an ordinary neural network. The flattened layer is passed to ReLU layer with 256 hidden neurons. This layer learns 29491456 parameters. Again one dropout layer is applied with 50\% threshold. In the last layer we use a softmax\cite{10.5555/2969830.2969856} activation output layer that gives the probabilities of both classes. Since this is a binary classification problem, we could've used sigmoid\cite{10.1016/S0893-6080(05)80129-7} activation, but we used softmax as we wanted to distinguish between the confidence attained in both the classes after prediction. The backward propagation of the network uses binary cross-entropy\cite{zhang2018generalized} as the loss function and Adam\cite{kingma2014adam} as loss optimizer function. The model has been trained for different set of epochs to observe the effect in different ways. One time for 50 epochs and other time for 100 epochs. These yield different observations which we will see in the Results section of this paper. 

\subsection{Results}
The result section gives how well the implementation has taken the lead. The model is defined by the 2 metrics, viz. Accuracy and Loss. Accuracy focuses on how well the model has learnt from the data with training set and validation set. Loss focuses on its reduction rate, closer to zero, better the loss. This is also calculated for training and validation.

\begin{figure}[htbp]
    \centering
    \includegraphics[scale=0.5]{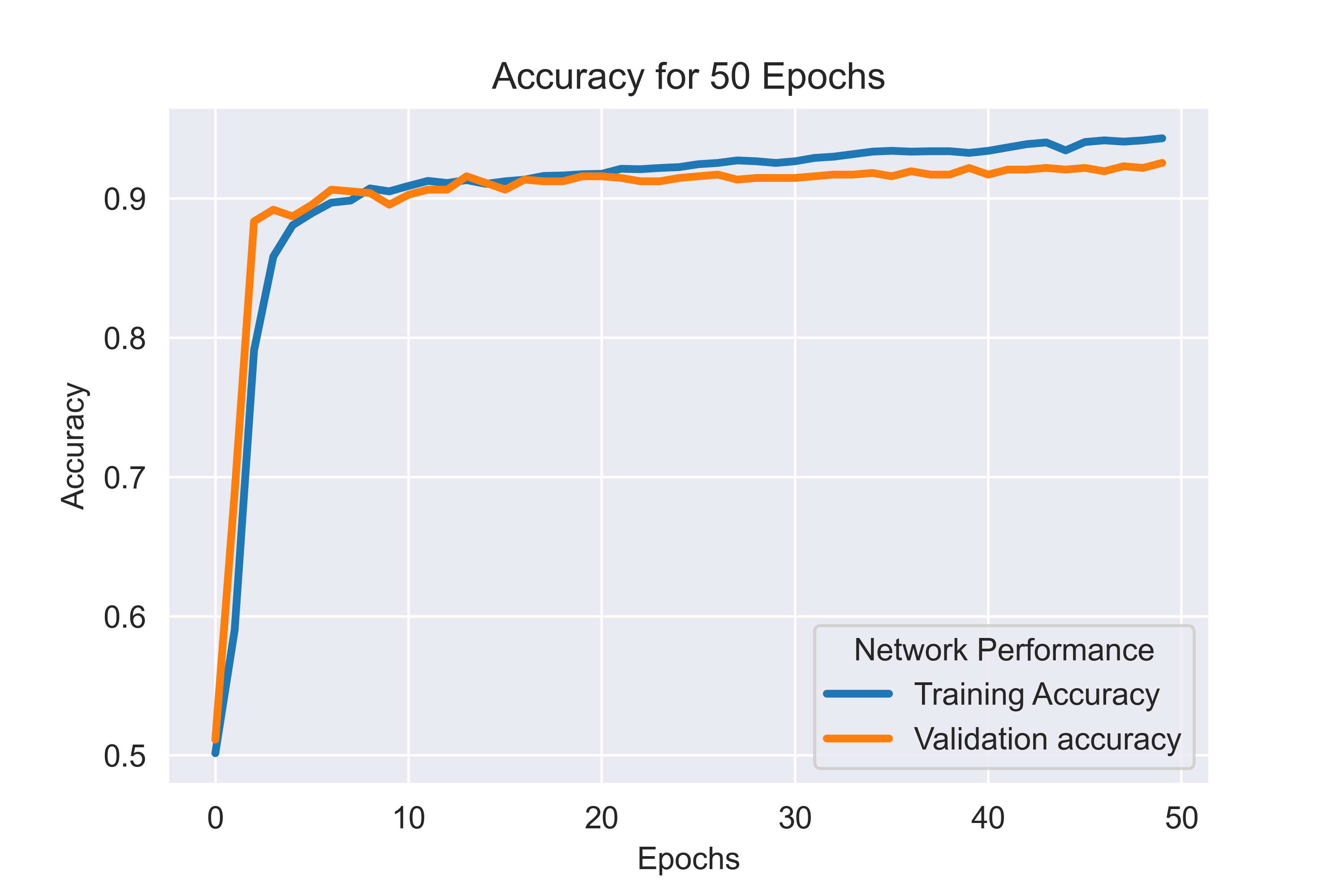}
    \caption{Accuracy for 50 Epochs}
    \label{fig:fig6}
\end{figure}

\begin{figure}[htbp]
    \centering
    \includegraphics[scale=0.5]{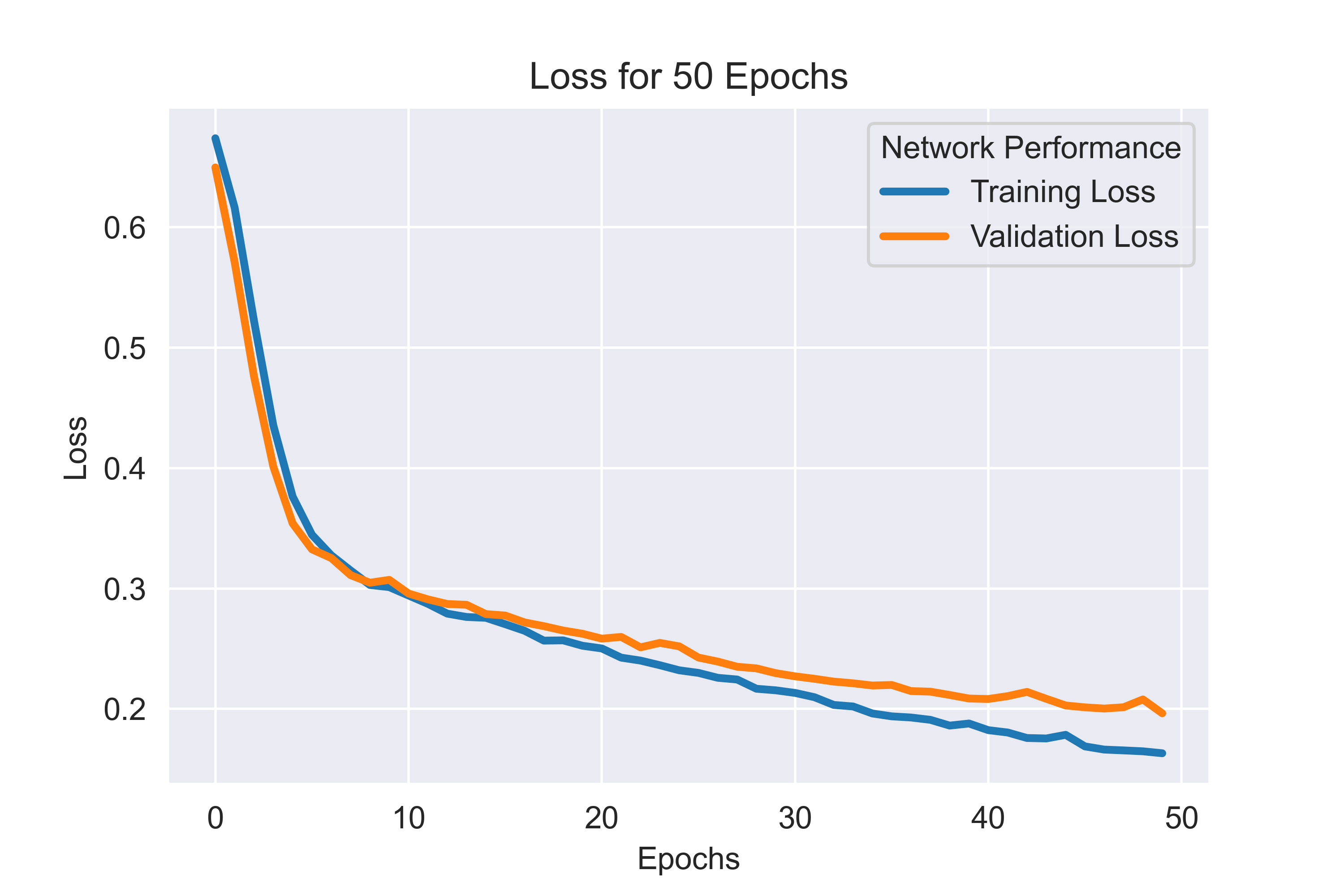}
    \caption{Loss for 50 Epochs}
    \label{fig:fig7}
\end{figure}

The accuracy for 50 epochs of training and validation can be seen in figure \ref{fig:fig6}. The exact numbers for training set is 94.33\% and validation set is 92.56\%. Similarly the loss values for training and validation set can be seen in figure \ref{fig:fig7}.

\begin{figure}[htbp]
    \centering
    \includegraphics[scale=0.55]{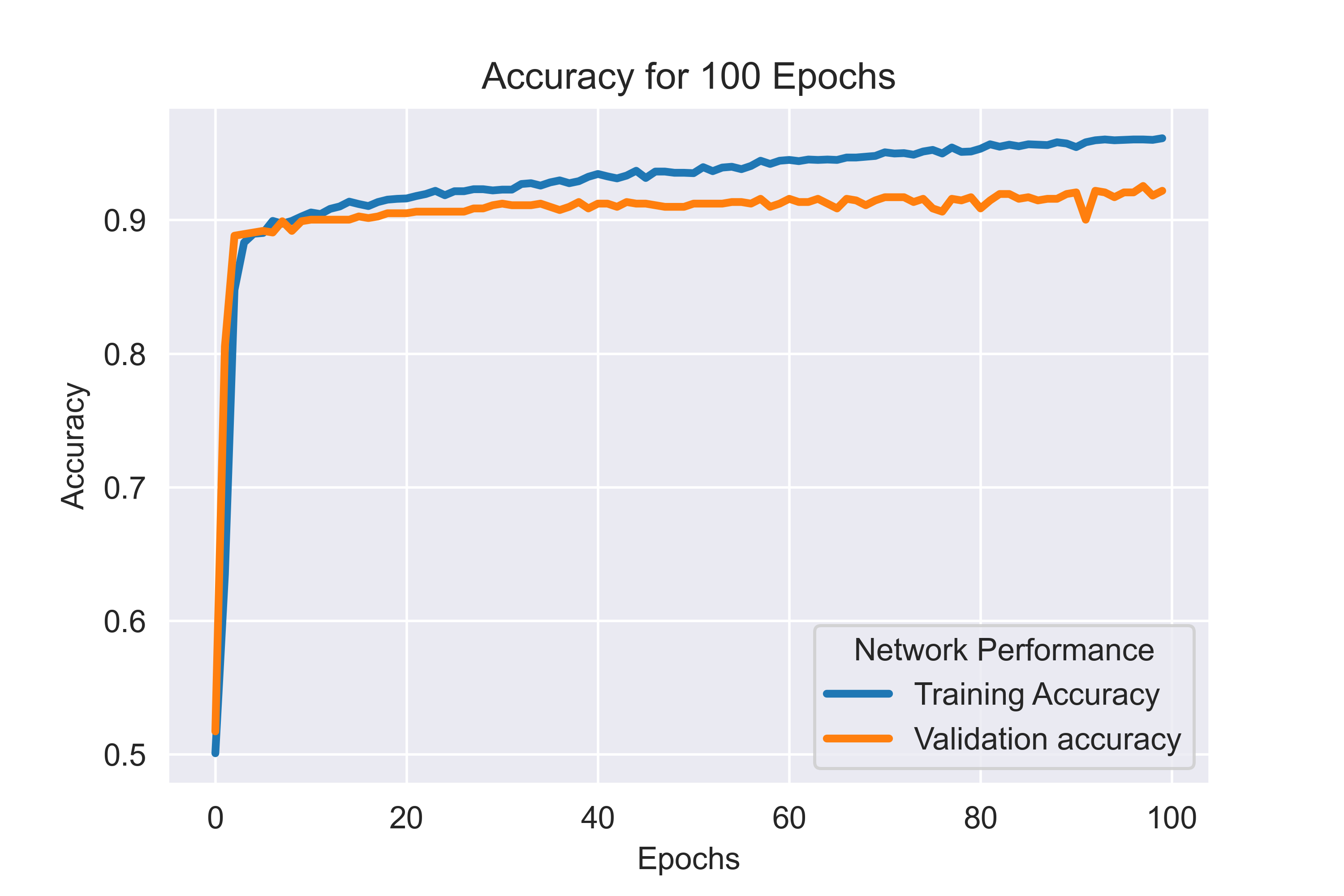}
    \caption{Accuracy for 100 Epochs}
    \label{fig:fig8}
\end{figure}

The exact values for the loss of training and validation for 50 epochs is 16.31\% and 19.63\%. Similar way of analysis for 100 epochs can be seen for training and validation accuracy in figure \ref{fig:fig8}.

\begin{figure}[htbp]
    \centering
    \includegraphics[scale=0.55]{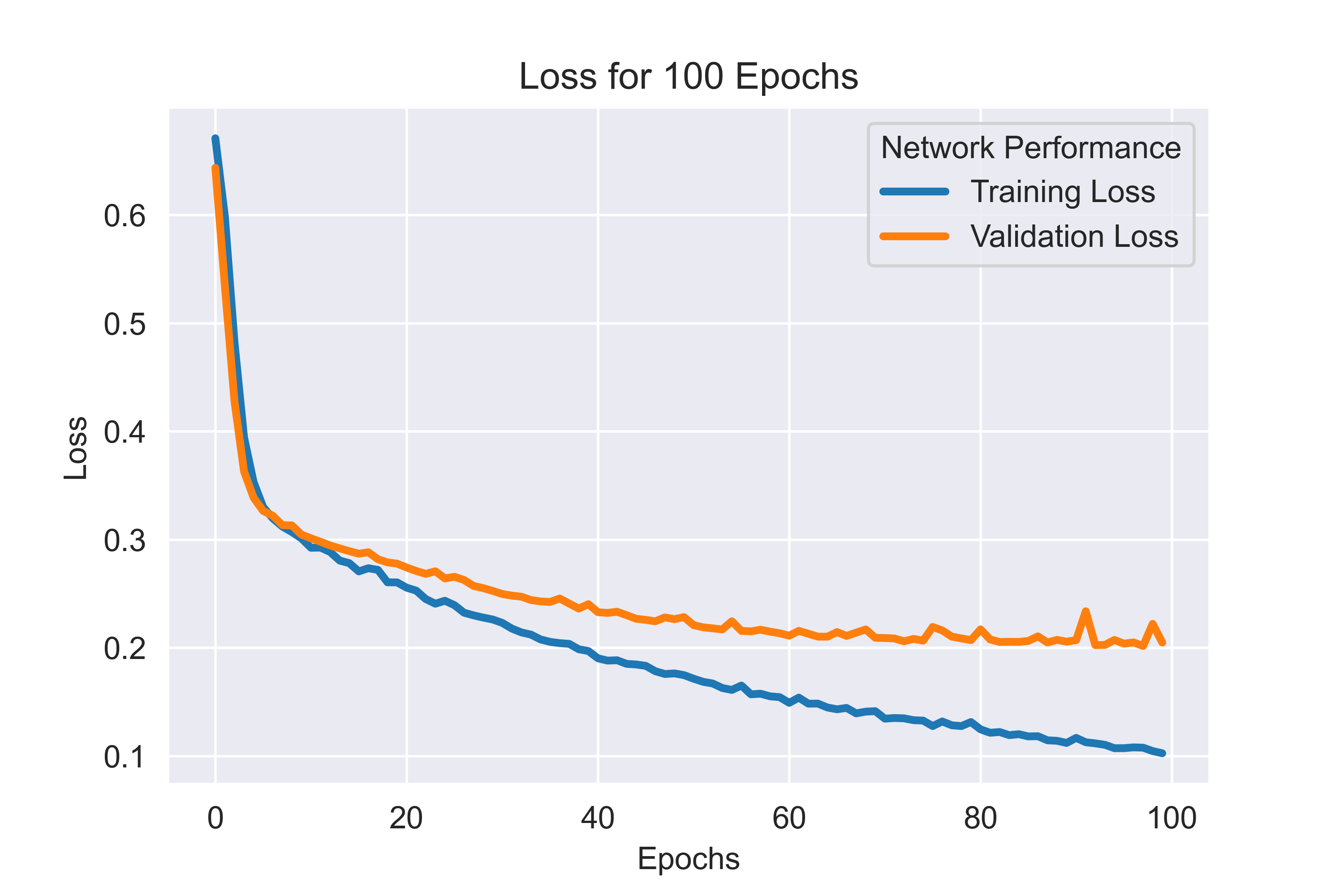}
    \caption{Loss for 100 Epochs}
    \label{fig:fig9}
\end{figure}

The exact values for the accuracy over 100 epochs with training and validation sets are 96.13\% and 92.20\% respectively. This same can be done for loss values over 100 epochs and can be seen in figure \ref{fig:fig9}.

The exact values for the loss over 100 epochs with training and validation sets are 10.27\% and 20.50\% respectively.

\section{Conclusion}
The tampering of images is done using image editing tools and this is a form of image forensics problem which in this paper is solved by leveraging deep learning. The Error Level Analysis technique is applied to every single image in the training data of CASIA that creates a good distinction between authentic and tampered images using editing tools. Later a simple Convolutional Neural Network is used to classify the 2 separate classes. Result section of the paper gives good insights on it and this is a very simple point we tried working with and later can be used to yield more effective applications for which we would be more than glad if this paper is used as a reference material.

%Bibliography
\bibliographystyle{unsrt}  
\bibliography{references}

\end{document}